\documentclass[letterpaper, 10 pt, conference]{ieeeconf}  

\IEEEoverridecommandlockouts                        
\title{\LARGE \bf
ManiLadder: Benchmarking Robot Manipulation Through a Categorized and Multi-Difficulty Level Task Ladder
}

\author{Chongkai Gao$^{*1}$, Jieao Shi$^{*1}$, Zhaoyu Mu$^{*1}$, Yudi Lin$^{*12}$,\\ Xuanwei Liu$^1$, Wei Ken Ng$^3$, Baiye Cheng$^4$, Anqi Chen$^5$, Kerun Liu$^6$, Zhuoran Li$^7$, Xuchuan Huang$^8$, \\ Haoyang Li$^7$
Shengjia Zhu$^7$, Nga Teng Chan$^9$, Yuxuan Zhang$^1$, Xingyu Chen$^1$, Lin Shao$^1$\\ \\$^1$National University of Singapore \quad $^2$Nanyang Technological University \quad $^3$University of Toronto \\ $^4$Huazhong University of Science and Technology \quad $^5$Nanjing University \quad $^6$KTH \\ 
$^7$Shanghai Jiao Tong University \quad $^8$Peking University \quad $^9$HKUST}%

\usepackage{bbm}
\usepackage{etoolbox}
\usepackage{multicol}
\usepackage{cite}
\usepackage[bookmarks=true]{hyperref}
\usepackage{graphics} 
\usepackage{times} 
\usepackage{amsmath} 
\usepackage{amssymb}  
\usepackage{svg}
\usepackage{mathrsfs}
\usepackage{amsfonts}
\usepackage{wrapfig}
\usepackage{amsmath}

\usepackage{kantlipsum}
\usepackage{subcaption}
\usepackage{marvosym}

\usepackage{graphicx}
\usepackage{float}
\usepackage{ctable}
\usepackage{cuted}
\usepackage{colortbl}
\usepackage{multirow}
\usepackage{wrapfig}
\usepackage[misc,geometry]{ifsym}
\usepackage{algorithm}
\usepackage{algpseudocode}
\usepackage{xcolor}   
\usepackage{pifont}    

\usepackage{bm}
\usepackage{caption}
\usepackage{setspace}
\let\labelindent\relax
\usepackage{enumitem}
\usepackage{threeparttable}
\newcommand{\TODO}[1][]{\textcolor{red}{\bf [TODO]}}
\setlist[enumerate,1]{itemsep=3pt}

\newcommand{\name}{ManiLadder}

\definecolor{formalgreen}{rgb}{0.1, 0.7, 0.1}  
\definecolor{formalred}{rgb}{0.9, 0.2, 0.2}  

\newcommand{\cmark}{\textcolor{formalgreen}{\checkmark}}  
\newcommand{\xmark}{\textcolor{formalred}{\ding{55}}}           

\usepackage{pgf}
\definecolor{findingframe}{HTML}{3C5488}
\newcounter{finding}
\newsavebox{\findingbox}
\newlength{\findingh}
\newcommand{\finding}[1]{%
  \par\addvspace{4pt}\refstepcounter{finding}%
  \sbox{\findingbox}{\parbox{\dimexpr\linewidth-14pt\relax}{\textbf{\textcolor{findingframe}{Finding~\thefinding:}} \textit{#1}}}%
  \setlength{\findingh}{\dimexpr\ht\findingbox+\dp\findingbox+10pt\relax}%
  \noindent\begin{pgfpicture}%
    \pgfsetcornersarced{\pgfpoint{4pt}{4pt}}%
    \pgfpathmoveto{\pgfpoint{0.5\linewidth}{0pt}}%
    \pgfpathlineto{\pgfpoint{\dimexpr\linewidth-0.6pt\relax}{0pt}}%
    \pgfpathlineto{\pgfpoint{\dimexpr\linewidth-0.6pt\relax}{\findingh}}%
    \pgfpathlineto{\pgfpoint{0pt}{\findingh}}%
    \pgfpathlineto{\pgfpoint{0pt}{0pt}}%
    \pgfpathclose
    \pgfsetfillcolor{findingframe!6}\pgfsetstrokecolor{findingframe}\pgfsetlinewidth{0.6pt}%
    \pgfusepath{fill,stroke}%
    \pgftext[left,bottom,at={\pgfpoint{7pt}{5pt}}]{\usebox{\findingbox}}%
  \end{pgfpicture}\par\addvspace{6pt}}

\begin{document}

\maketitle
\thispagestyle{empty}
\pagestyle{empty}

\begin{strip}
\vspace{-0.3in}
\centering
\includegraphics[width=\linewidth]{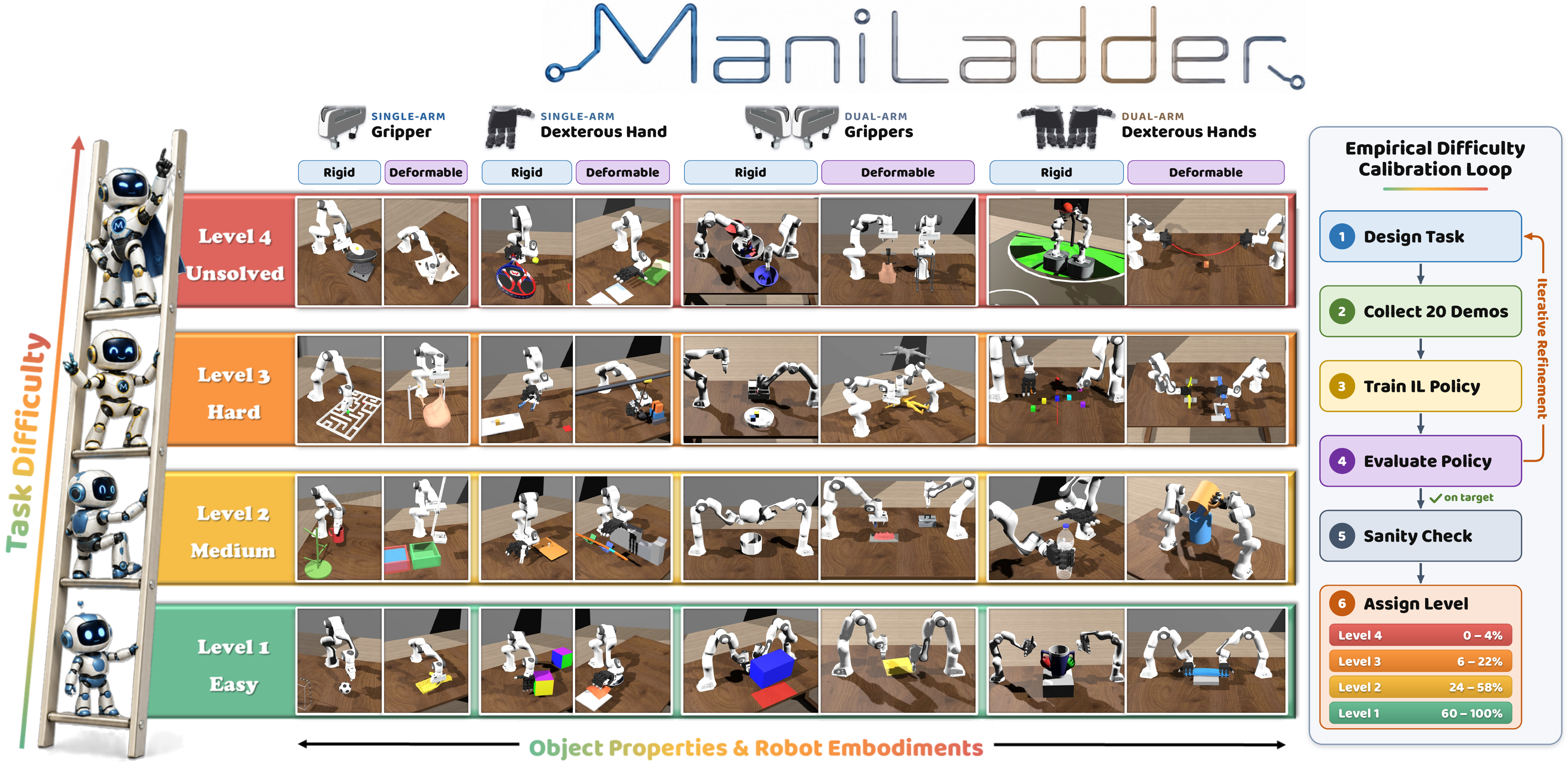}
\captionof{figure}{Overview of \name, a difficulty-calibrated simulation benchmark for robotic manipulation. \name\ organizes 112 tasks (only 32 are shown here) into a vertical ladder of four difficulty levels while maintaining broad horizontal diversity across robot embodiments, object properties, and skills. The difficulty of each task is calibrated and iteratively refined by the success rate of a standardized imitation-learning policy and later verified under different seeds and policies, enabling evaluation of both whether a model can solve a task and the level of manipulation difficulty it can reliably reach.}
\label{fig:intro}
\end{strip}


\begin{abstract}
Recent advances in robotic manipulation have enabled increasingly complex and dexterous behaviors, yet existing benchmarks still focus on simple pick-and-place tasks and lack a systematic framework for measuring progress in robot learning along the axis of task difficulty. We present \name, a large-scale and organized simulation benchmark for robotic manipulation. \name\ contains 112 carefully designed tasks stratified into 4 difficulty levels, with 8 categories in each level, spanning rigid-body and deformable-object manipulation, single-arm and dual-arm settings, and gripper and dexterous-hand end-effectors. To calibrate difficulty, we fix the data amount the same for every task and use a unified flow-matching imitation learning policy under the same training and evaluation protocols, then use its evaluation success rate as an operational empirical difficulty signal. All tasks are iteratively refined until they fall into target difficulty ranges, which are then verified across different random seeds, policies, and agentic learning systems. We further analyze the task properties, compare different policy performances and training settings, analyze task transfer and data scaling behaviors, and conduct real-world experiments showing that the designed tasks preserve their difficulty ordering beyond simulation.  \name\ provides a calibrated ladder for measuring, comparing, and advancing robotic manipulation capabilities. More results are on   \href{https://maniladder-benchmark.github.io/}{https://maniladder-benchmark.github.io/}.

\end{abstract}
\section{Introduction}

\begin{table*}[t]
    \centering
    \caption{Comparison between \name\ and other simulation robot manipulation benchmarks focusing on task taxonomy and difficulty levels. Auto Demo means supporting automatic data generation. We use $\mathcal{G}$ for robots with grippers, $\mathcal{X}$ for robots with dexterous hands, $\mathcal{S}$ for single-arm robots, $\mathcal{D}$ for dual-arm robots, and $\mathcal{M}$ for mobile robots.}
    \vspace{-0.05in}
    \resizebox{\textwidth}{!}{
    \begin{tabular}{c|cccccc}
    \toprule
    Benchmark & Simulator & No. of tasks & Auto Demo & Difficulty Taxonomy & Deformable objects & Robots\\
    \midrule
    MetaWorld~\cite{metaworld}  & Mujoco~\cite{todorov2012mujoco} & 50 & \cmark & \xmark & \xmark & $\mathcal{S},\mathcal{G}$\\
    RoboDojo~\cite{chen2026robodojo} & Isaac Sim~\cite{isaaclab} & 42 & \cmark & \xmark & \cmark & $\mathcal{D},\mathcal{G}$\\
    COLOSSEUM~\cite{colosseum} & V-Rep~\cite{pyrep} & 20 & \cmark & \xmark & \xmark & $\mathcal{S},\mathcal{G}$\\
    RoboCasa~\cite{robocasa} & Mujoco~\cite{todorov2012mujoco} & 100 & \cmark & \xmark & \xmark & $\mathcal{S},\mathcal{X},\mathcal{G},\mathcal{D},\mathcal{M}$\\
    Behavior 1K~\cite{behavior1k} & Isaac Sim~\cite{isaaclab} & 1000 (100 released) & \xmark & \xmark & \cmark & $\mathcal{S},\mathcal{G},\mathcal{M}$\\    
    LIBERO~\cite{libero} & Mujoco~\cite{todorov2012mujoco} & 120 & \xmark & \cmark & \xmark & $\mathcal{S},\mathcal{G}$\\
    RoboLab~\cite{robolab} & Isaac Sim~\cite{isaaclab} & 120 & \cmark & \cmark & \xmark & $\mathcal{S},\mathcal{G}$\\    
    \textbf{\name\ (ours)} & Genesis~\cite{genesis} & 112 & \cmark & \cmark & \cmark & $\mathcal{S},\mathcal{X},\mathcal{G},\mathcal{D},\mathcal{M}$\\
    \bottomrule
    \end{tabular}}
    \vspace{-0.15in}
    \label{tab:benchmark_comparision}
\end{table*}

Recent years have witnessed remarkable progress in robotic manipulation, driven by increasingly accessible hardware~\cite{act,leaphand}, large-scale datasets~\cite{oxe,chen2026objectworld}, powerful foundation models~\cite{pi05}, effective learning algorithms~\cite{dp}, and agentic learning systems~\cite{rax}. Robots can now perform highly complex real-world tasks such as scrambling eggs~\cite{skild_ai}, tying plastic bags~\cite{gao2023iterative}, and cardboard box folding~\cite{dreamcan}. However, current benchmarks in robotic manipulation lag considerably behind this huge progress: most existing robotic simulation benchmarks~\cite{libero,robocasa,behavior1k} primarily expand task diversity \textit{horizontally}, by recombining relatively simple manipulation skills across different scenes and objects, such as pick-and-place, pushing, and door opening. While they are useful for evaluating generalization or semantic reasoning, they offer limited resolution for measuring the \textit{vertical} growth of manipulation capability, i.e., they cannot reveal whether a model has truly expanded the frontier of solvable manipulation tasks. We therefore ask a capability-oriented question: \textbf{what level of manipulation difficulty can a model reliably solve?} Answering this requires a benchmark organized by calibrated difficulty levels, so that the capability boundaries of current methods can be measured directly.




Building such a benchmark raises two challenges: (1) how to determine task difficulty, and (2) how to construct diverse tasks with well-calibrated difficulty. Prior works characterize difficulty through theoretical measures~\cite{informationinvariant}, subskill composition~\cite{dmil}, manually defined predicates~\cite{tamp}, or contact patterns~\cite{classifying}, but these abstractions are difficult to scale to diverse modern manipulation tasks. Instead, we adopt an empirical definition: under a standardized training protocol, what success rate can a robot learning algorithm achieve on the task? This becomes increasingly practical as modern imitation learning methods~\cite{flowmatching,act} perform effectively across a broad range of tasks. Meanwhile, although modern simulators and procedural generation tools~\cite{genesis,isaaclab} simplify task creation, they do not guarantee task difficulty. Initialization ranges, perception complexity, and object geometry can make tasks unintentionally too easy or too hard. Constructing a meaningful difficulty ladder therefore requires iterative, metric-driven task calibration.

In this work, we present \name, a large-scale simulation benchmark that evaluates robotic manipulation along two complementary axes: calibrated task difficulty and manipulation diversity, as shown in Figure~\ref{fig:intro}. \name\ contains 112 carefully designed tasks across 4 difficulty levels, organized by object properties, robot embodiments, and manipulation skills. The tasks cover both rigid-body and deformable-object manipulation, spanning four embodiment types: single-arm grippers, single-arm dexterous hands, dual-arm grippers, and dual-arm dexterous hands. Within each task suite, we further cover disparate manipulation skills or object types, resulting in 3 to 4 concrete tasks per suite. For example, in each deformable-object manipulation suite, we design tasks targeting 1D (e.g., rope), 2D (e.g., cloth), and 3D (e.g., clay) deformable object types, respectively. To stratify task difficulty, we provide a fixed number of demonstrations for every task and train a unified flow-matching~\cite{flowmatching} imitation learning policy under a standardized protocol, using its evaluation success rate as an empirical difficulty signal. We then iteratively calibrate each task to the desired difficulty range by adjusting interpretable factors such as object geometry, initial randomization, perceptual difficulty, temporal coordination, and error tolerance. The calibrated task difficulties are then verified across demonstration recollection, retraining, different policies, and agentic learning systems. Building on these calibrated difficulty levels, we establish standardized benchmarking protocols and an Attained Level metric to quantify the highest task difficulty level a given algorithm can reliably solve. This makes \name{} not only broad in task coverage but also explicitly calibrated as a ladder of manipulation difficulty.

To highlight the flexibility and potential of \name\ for studying manipulation capability across different dimensions, we conduct a set of studies, including: 1) analyzing the task properties across different levels; 2) comparing small and large VLA models~\cite{pi05} on \name; 3) studying the multi-task training results compared to single-task training results; 4) studying task transfer ability and data scaling behaviors across different task families; 5) examining whether the difficulty ordering calibrated in simulation is preserved between simulation and real-world settings. Together, \name\ provides a platform for studying many future questions in robot learning, and also provides a calibrated and organized difficulty ladder that measures how far robotic manipulation has climbed toward general-purpose manipulation intelligence, and how far it still has to go.

\section{Related Works}
\subsection{Robot Manipulation Benchmarks}

Robot manipulation benchmarks span simulation and real-world settings. Simulation benchmarks can offer diverse tasks~\cite{rlbench,robosuite} and study of long-horizon execution~\cite{calvin,robocasa}, multi-task, transfer, and continual learning~\cite{libero,metaworld,cril}, robustness~\cite{colosseum}, and dexterous~\cite{dexart} or bimanual manipulation~\cite{chen2025robotwin}. Real-world evaluation efforts promote comparability through standardized objects~\cite{ycb}, assembly protocols~\cite{furniturebench}, distributed policy comparisons~\cite{roboarena}, and in-person challenges~\cite{amazonchallenge}. RoboDojo~\cite{chen2026robodojo} further connects simulation and real-world evaluation. Simulation enables controlled, scalable, and reproducible evaluation, while real-world benchmarks directly assess performance under physical deployment conditions. We develop \name{} primarily in simulation and focus on calibrating difficulty across diverse tasks through controlled configurations and repeated evaluation. We further evaluate selected real-world counterparts across task categories, providing a preliminary assessment of whether the relative difficulty trends persist in physical execution.

\subsection{Robot Manipulation Task Taxonomies}

Developing a systematic taxonomy for robot manipulation is inherently challenging because task complexity arises jointly from object properties, interaction modes, embodiment, and temporal structure~\cite{classifying}. Existing taxonomies have been developed based on grasp and contact modes~\cite{cutkosky1989grasp}, predefined skills or primitives~\cite{huang2021robotic}, and specialized domains such as deformable~\cite{tdom,flip,chen2025metafold} or bimanual manipulation~\cite{bimanual}. While useful for describing manipulation structure, these taxonomies are often tied to specific interaction patterns or skill decompositions, making them difficult to apply consistently to heterogeneous, multi-stage tasks, and unsuitable for studying task difficulties.


We instead organize manipulation tasks along two complementary axes: horizontally by task diversity, primarily characterized by manipulated objects and robot embodiments, and vertically by empirical difficulty, measured by success rates under a unified policy, training, and evaluation protocol. This provides a common behavioral scale for comparing heterogeneous manipulation tasks without relying on hand-crafted notions of complexity.

\section{\name{} Task Construction and Difficulty Calibration}
\label{sec:construction}

\begin{table*}[t]
    \centering
    \caption{
    Manipulation skill coverage of \name. 
    We group skills according to prior literature on manipulation-primitives 
    and contact-rich manipulation taxonomy~\cite{huang2023autogenerated,morrow1997manipulation,zech2019action,feix2016grasp}.
    }
    \label{tab:skill_coverage}
    \resizebox{\textwidth}{!}{
    \begin{tabular}{l l c c}
        \toprule
        \textbf{Skill family} 
        & \textbf{Primitive skills} 
        & \textbf{\# Tasks} & \textbf{Avg. Level} \\
        \midrule
        Prehensile manipulation 
        & grasp, pick, hold, lift, carry, place, release 
        & 33 & 2.36 \\
        
        Non-prehensile manipulation 
        & push, pull, slide, drag, shift, nudge, pivot, tilt, flip
        & 9 & 2.22 \\
        
        Articulated-object manipulation 
        & open, close, turn, twist, press, pull-handle, push-button 
        & 16 & 2.07 \\
        
        Contact-rich assembly 
        & insert, extract, align, slide-along-contact, move-until-contact 
        & 17 & 2.88 \\
        
        Tool-use manipulation 
        & scoop, sweep, wipe, hook, lever, strike, stir 
        & 15 & 3.20 \\
        
        Deformable-object manipulation 
        & fold, unfold, hang, stretch, wrap, squeeze, flatten 
        & 48 & 2.50 \\
        
        Granular and fluid manipulation 
        & pour, fill, transfer, scoop, spill-control, measure 
        & 16 & 2.50 \\
        
        Bimanual coordination 
        & handover, stabilize-and-manipulate, dual-arm lift \& pull \& push 
        & 56 & 2.50 \\
        
        Dynamic manipulation 
        & drop, toss, throw, catch, bounce, hit, swing 
        & 15 & 3.47\\
        \bottomrule
    \end{tabular}
    }
\end{table*}

In this section, we describe how \name\ is constructed as a difficulty-calibrated simulation benchmark for robotic manipulation across diverse task categories. We first introduce the design principles and detailed task construction process, then we introduce how to iteratively refine tasks to calibrate their difficulty levels. This leads to the final standardized \name{} benchmark, which will be introduced in the next Section.

\subsection{Design Principles and Task Organization}
\label{sec:construction:organization}

\name\ is designed to have four difficulty levels and to measure the level of manipulation difficulty that a robot learning policy can reliably solve. To make such a comparison convincing, tasks in each level must be diverse enough as well as organized in a structured way, so that the performance differences mostly reflect changes associated with task difficulty. We design \name\ along three axes: all levels contain unified task suites organized by object property, robot embodiment, and manipulation skill, as shown in Figure \ref{fig:intro}.

\begin{figure}
    \centering
    \includegraphics[width=\linewidth]{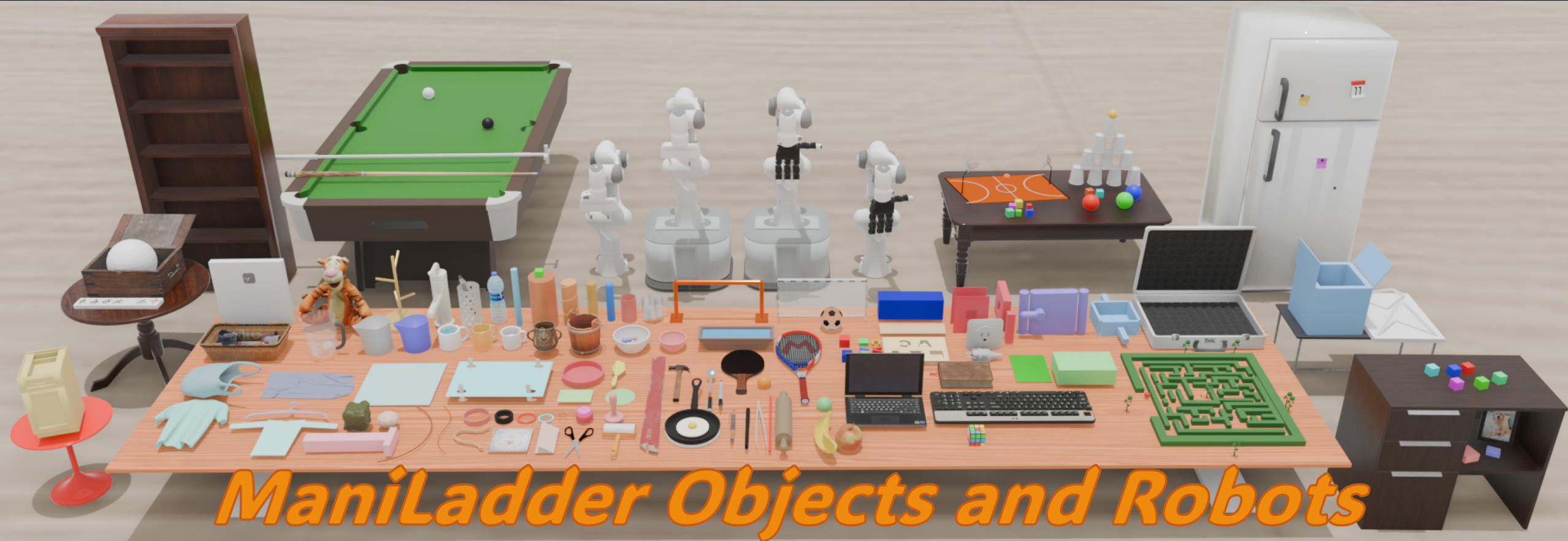}
    \caption{The objects and robots in \name. Four kinds of robots and diverse kinds of rigid-body and deformable objects are included in \name.}
    \vspace{-0.1in}
    \label{fig:scene}
\end{figure}

The first design axis is \textbf{object physical property}. We divide tasks in each level into rigid-body manipulation and deformable-object manipulation. Rigid-body tasks include both objects with fixed geometry, such as blocks and mugs, and articulated objects with movable parts, such as doors and drawers. Deformable-object tasks are grouped by geometric dimensionality: 1D deformables such as ropes and cables, 2D deformables such as cloths and towels, and 3D deformables such as clay, plasticine, and liquid. This distinction allows ManiLadder to cover qualitatively different interaction regimes. We curate object assets from multiple open-source repositories~\cite{ycb} and custom designs to support broad task coverage. These assets are rescaled, cleaned, and processed with convex decomposition to improve simulation stability. For deformable-object manipulation, we create assets in Blender and convert them into simulation-ready formats with Trimesh, and use different physical representations for different object categories: 1D ropes are represented as articulated chains of capsule-shaped rigid bodies, 2D cloths are simulated with a PBD solver, and 3D deformable materials are modeled with solvers such as MPM for clay-like objects and SPH for liquids. Together, these assets allow \name\ to cover a wide range of manipulation regimes, from rigid-body interaction to contact-rich deformable-object manipulation, as shown in Figure~\ref{fig:scene}. To the best of our knowledge, \name\ is the first simulation benchmark that contains more than 40 deformable object manipulation tasks within a well-organized structure.

\begin{figure*}
    \centering
    \includegraphics[width=\linewidth]{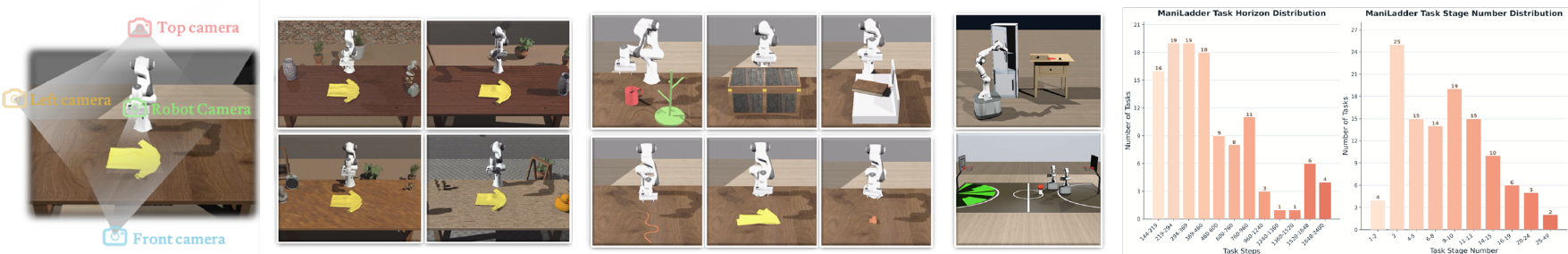}
    \caption{Left: the basic scene setup of task \texttt{FoldShirt}, serving as an example illustrating a typical task scenario in \name. Middle left: the domain randomization on the lighting, background, and distracting objects of \texttt{FoldShirt}. Middle: in each rigid-body object task suite, we design different tasks focusing on different manipulation skills, such as \texttt{HangMug} for pick-and-place, \texttt{OpenBoxRetrival} for articulated object manipulation, and \texttt{PivotBook} for non-prehensile manipulation. In each deformable object task suite, we design tasks including 1D, 2D, and 3D deformable objects, such as \texttt{StraightenRope} for rope manipulation, \texttt{UnfoldShirt} for cloth manipulation, and \texttt{ShapeTtoI} for clay manipulation. Middle right: two mobile manipulation tasks in \name. Right: the task horizon and stage number histograms.}
    \label{fig:domain_random}
\end{figure*}

The second design axis is \textbf{robot embodiment}. We consider four end-effector configurations: single-arm gripper, single-arm dexterous hand, dual-arm gripper, and dual-arm dexterous hand. These embodiments introduce different manipulation capabilities and coordination requirements, ranging from prehensile grasping to dexterous in-hand and bimanual coordination. We use Franka Panda as the robot arm and LeapHand~\cite{leaphand} as the dexterous hand. A small number of tasks further include mobile manipulation when base motion is intrinsic to the task, using TidyBot++~\cite{tidybot++}.

The third design axis is \textbf{manipulation skill}. Previous simulation benchmarks mainly focus on simple pick-and-place or push tasks. In \name{}, we aim to significantly expand the scope of skill coverage. Within each object-property and embodiment category, we instantiate three to four tasks that cover distinct skills whenever possible, as shown in the middle of Figure~\ref{fig:domain_random}. Our skill selection is a comprehensive consideration and summary of prior studies on manipulation primitives, contact-based manipulation behaviors, action representations, and grasp taxonomies~\cite{tamp}. Table~\ref{tab:skill_coverage} summarizes the manipulation skills and their corresponding number of occurrences and levels in \name. This design avoids concentrating a task suite around a single operation and encourages each difficulty level to cover multiple sources of manipulation challenge. Overall, \name\ contains 112 manually curated tasks arranged in an aligned hierarchy of difficulty level, object property, embodiment, and skill.

\name\ is built upon Genesis~\cite{genesis}, a cross-platform simulator that supports rigid-body dynamics, deformable objects, and GPU-based parallel simulation. We deliberately focus on manipulation-centric challenges rather than scene-level diversity or navigation complexity. Therefore, most tasks are instantiated in controlled tabletop environments, with a small number of mobile manipulation scenarios included only when base motion is intrinsic to the task. We implement a unified simulation interface following the Gym-style API across all tasks, standardizing the policy-environment interaction protocol, including environment reset, observation retrieval, action execution, reward computation, success checking, and evaluation rollout. We implement a scripted policy for each task to synthesize demonstrations in the simulation. We set three third-view cameras and one or two first-view cameras on robots, and add domain randomization in lighting conditions, table and background textures, and distracting objects.

\begin{figure*}
    \centering
    \includegraphics[width=\linewidth]{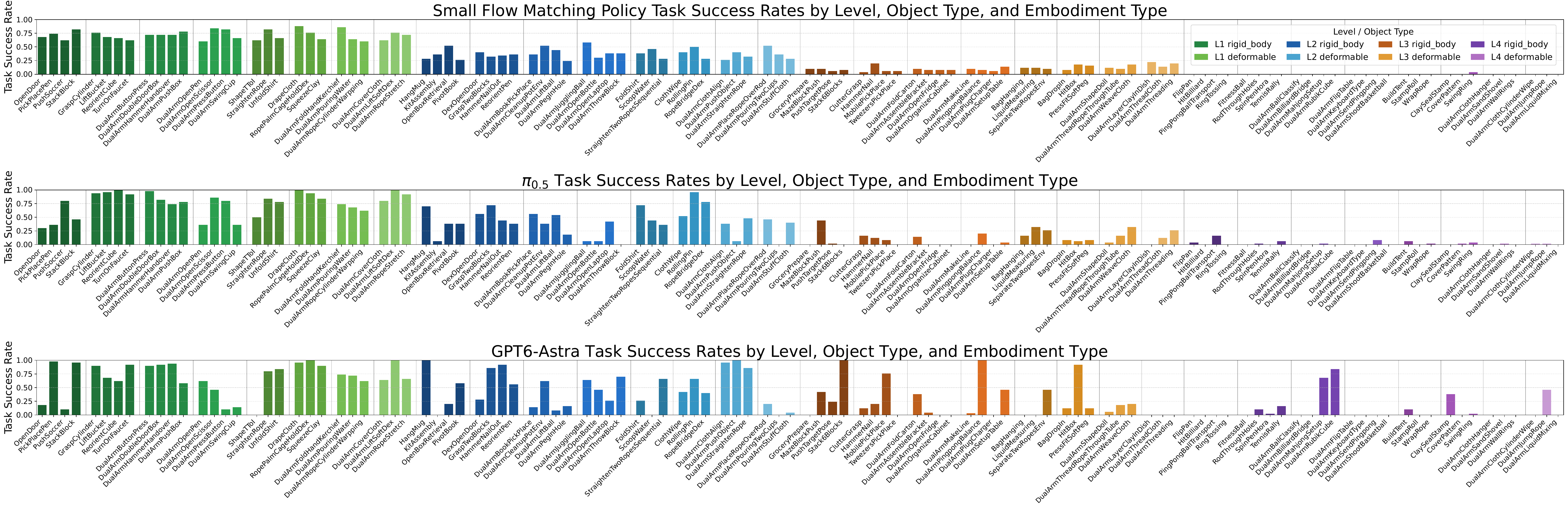}
    \caption{All task success rates in \name\ of the small flow-matching policy, $\pi_{0.5}$, and GPT6-Astra agentic script policy.}
    \label{fig:all_success_rates}
\end{figure*}

\subsection{Task Difficulty Calibration and Iterative Refinement}

Through the above design process, we create 112 simulation manipulation tasks. Now we need to calibrate and refine the difficulty of each task to make sure it falls into the correct difficulty level. We acquire task difficulty signals operationally using a standardized reference imitation learning policy: under the same number of demonstrations, model architecture, hyperparameters, training budget, and evaluation protocol, harder tasks are those on which the reference policy achieves lower success rates.  We emphasize that \name\ does not aim to define the real and absolute, policy-independent task difficulty. Instead, for modern robot learning benchmarks, the practical question is often not whether a task is intrinsically difficult, but whether a policy class can reliably solve the task from a standardized amount of demonstration data and training budget. Under this view, a good and widely used reference policy could serve as a calibrated measuring instrument.

\begin{figure}
    \centering
    \includegraphics[width=\linewidth]{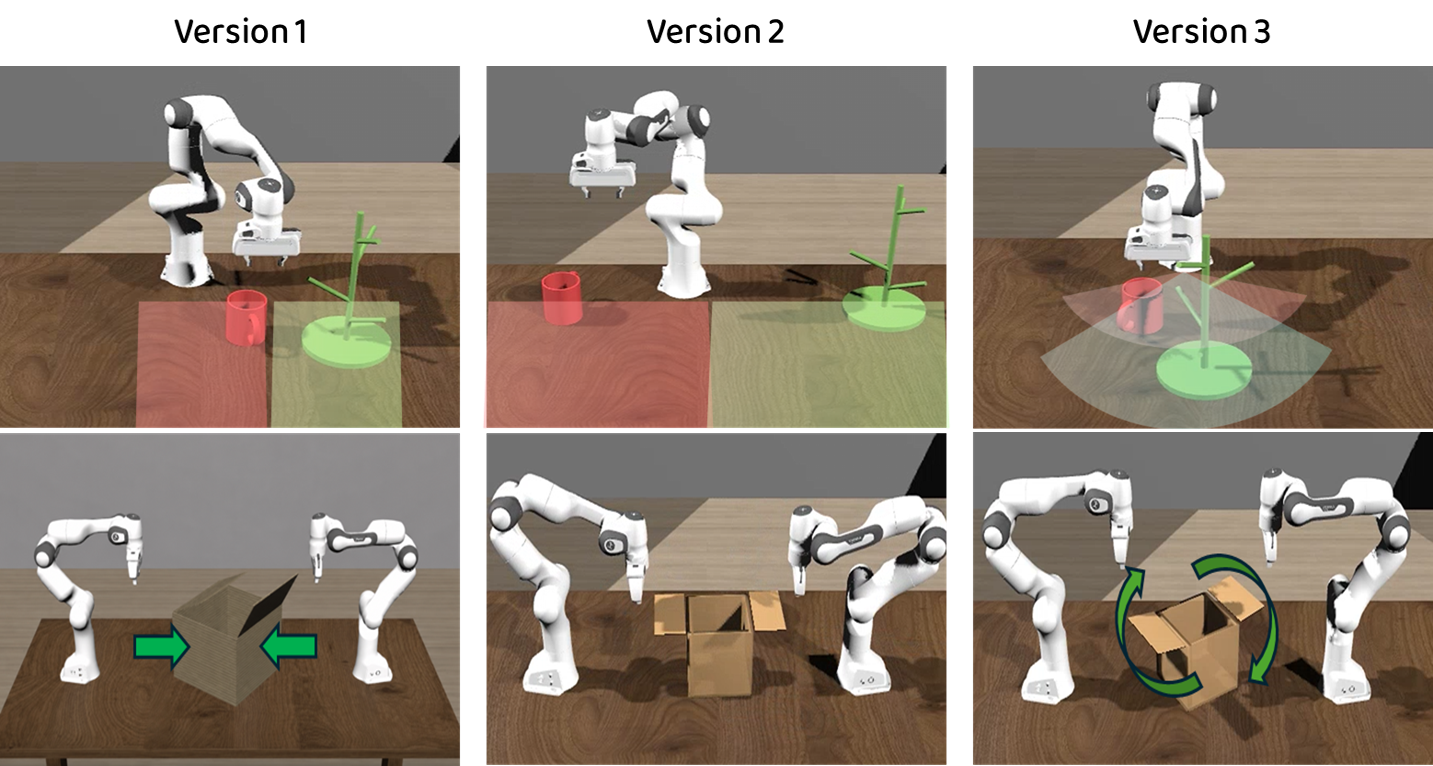}
    \caption{Examples of task calibration and refinement. Top: the \texttt{HangMug} task: we progressively modify the initialization regions of the mug and rack to make the task more challenging. Bottom: the \texttt{DualArmCloseBox} task: we first reduce the box size, then introduce an initialization pose that requires coordinated dual-arm rotation before the box can be closed.}
    \label{fig:refinement}
\end{figure}

We calibrate each task difficulty using a small standard flow-matching imitation learning policy with 50M learnable parameters, with RADIO~\cite{radio} as the pretrained frozen visual encoder for multi-view RGB image input, and a transformer encoder-decoder architecture for flow-matching~\cite{flowmatching} action chunk denoising. For each task $T$, we give 20 demonstrations and train the policy for $40k$ steps with a batch size of $256$. We evaluate every $5k$ steps for $50$ episodes and calculate the task success rate. We use the best evaluation success rate during training and assign tasks to four levels according to:
\begin{flalign*}
\ell(T)=\mathrm{Level}(T) =
\begin{cases}
1, & s(T) \in [60\%, 100\%], \\
2, & s(T) \in [24\%, 58\%], \\
3, & s(T) \in [6\%, 22\%], \\
4, & s(T) \in [0\%, 4\%].
\end{cases}
&&
\end{flalign*}
Here, Level 1 contains tasks that are reliably learnable by the reference policy, while Level 4 contains tasks that remain very hard under the same learning budget.

If a task does not fall into its intended difficulty interval, we iteratively refine it by modifying the task design, recollecting demonstrations, retraining the reference policy, and reevaluating its success rate, as illustrated in the right part of Figure~\ref{fig:intro}. For example, if a Level-2 task has a success rate of $30\%$, we need to modify the task to decrease its difficulty. During this process, we adjust task difficulty through several human-interpretable factors, such as task horizon, spatial error tolerance, object complexity, and spatial-temporal coordination. Examples are shown in Figure~\ref{fig:refinement}. On average, it requires about four rounds of such a refinement process to make a task calibrated, which shows a large gap between subjective human intuition and empirical difficulty signal under a standardized reference protocol. 

Although the aforementioned calibration process is objective, it remains subject to limitations regarding the number of evaluation trials and the specific policy employed; for instance, results may vary depending on the random seeds used for demonstration collection and evaluation, or different policy architectures applied to the same task might yield different difficulty levels. Consequently, after refining the tasks using this calibration process, we conducted additional validation in two directions:

1) \textbf{Random seed experiments}. Under three different random seeds, we re-collect 20 demonstrations for each task, followed by re-training, re-evaluation, and re-calculating the peak success rate. Let $\mathcal{T}$ denote the fixed set of $N=112$ tasks.
For each task $T$, let $s_T^{(0)} \in [0,1]$ denote its
original calibration score and $s_T^{(r)}$ its score in
repetition $r \in \{1,\ldots,R\}$, where $R=3$.
Applying the same prespecified thresholds yields the
corresponding levels $\ell_T^{(0)}$ and $\ell_T^{(r)}$. We assess reproducibility using three complementary metrics.
Let $\mathbf{q}^{(r)}$ be the vector of task ranks induced
by $\{s_T^{(r)}\}_{T\in\mathcal{T}}$, assigning average
ranks to tied scores. We define:
\begin{equation}
\begin{aligned}
    \rho_r
    &=
    \operatorname{Corr}
    \left(\mathbf{q}^{(0)},\mathbf{q}^{(r)}\right), \\
    \Delta_r
    &=
    \frac{100}{N}
    \sum_{T\in\mathcal{T}}
    \left|s_T^{(r)}-s_T^{(0)}\right|, \\
    A_r
    &=
    \frac{100}{N}
    \sum_{T\in\mathcal{T}}
    \mathbf{1}
    \left[\ell_T^{(r)}=\ell_T^{(0)}\right],
\end{aligned}
\label{eq:calibration_reproducibility}
\end{equation}
where $\operatorname{Corr}$ denotes Pearson correlation
between the rank vectors.
Spearman correlation $\rho_r$ measures agreement in task
ordering; mean absolute success-rate change $\Delta_r$
measures score variation; and level
agreement $A_r$ measures the percentage of tasks retaining
their original levels.

\begin{table}[t]
    \centering
    \caption{
    Reproducibility of difficulty calibration under
    demonstration recollection and retraining, and two new calibration policies.
    Each metric is computed over all $112$ tasks.
    }
    \label{tab:calibration_reproducibility}
    \small
    \setlength{\tabcolsep}{4pt}
    \begin{tabular}{@{}lccc@{}}
        \toprule
        Setting
        & $\rho$ $\uparrow$
        & $\Delta$ (pp) $\downarrow$
        & $A$ (\%) $\uparrow$ \\
        \midrule
        3 random seeds
        & $0.958\pm\text{0.006}$
        & $\text{5.67}\pm\text{0.361}$
        & $\text{96.4}\pm\text{3.100}$ \\
        $\pi_{0.5}$
        & $\text{0.828}$
        & $\text{12.95}$
        & $\text{73.3}$ \\        
        GPT6-Astra
        & $\text{0.605}$
        & $\text{18.26}$
        & $\text{60.9}$ \\       
        \bottomrule
    \end{tabular}
    \vspace{-0.1in}
\end{table}

2) \textbf{Different calibration policy}. We additionally perform the same calibration using $\pi_{0.5}$~\cite{pi05}, a large pre-trained flow-matching VLA model, and the GPT6-Astra script policy, an agentic policy where we use GPT-6 Astra Medium as a task-specific coding agent to synthesize executable Python policies from 20 expert demonstrations and up to 20 environment interaction episodes. The agent analyzes RGB observations, robot proprioception, and demonstrated actions, then iteratively refines and independently rehearses script policies. Here we do not modify the task anymore.

The first row of Table~\ref{tab:calibration_reproducibility} shows that
independently recollecting demonstrations and repeating
training and evaluation yield high task-level rank
correlation with the original calibration and small
absolute changes in success rate. This demonstrates the robustness of the task difficulty levels in \name\ across different random seeds.

For the comparison between small flow matching policy, $\pi_{0.5}$, and GPT6-Astra policy, the second row and third row in Table~\ref{tab:calibration_reproducibility} show that the task rankings between different policies show strong correlation ($0.828$ between the small flow-matching policy and $\pi_{0.5}$, and $0.605$ between the small flow-matching policy and the GPT6-Astra policy), with more than half of the tasks ($73.3\%$ for $\pi_{0.5}$ and $60.9$ for GPT6-Astra policy) maintaining the same difficulty level. From Figure~\ref{fig:all_success_rates}, we can also see that average success rates decrease monotonically from Level 1 to Level 4 across all three methods, supporting the transferability of \name's calibrated
difficulty hierarchy beyond the reference flow-matching policy. 

However, this agreement at the level of task groups coexists with substantial task-specific differences. The small flow-matching policy and $\pi_{0.5}$ exhibit relatively similar performance profiles, with $\pi_{0.5}$ achieving higher average success rates at every level. GPT6-Astra shows a more distinct profile: it achieves similar results to both imitation-learning policies on Level 1, but achieves higher average success rates on Levels 2–4. Its gains are particularly pronounced on tasks such as \texttt{Stack6Blocks-v1} and \texttt{DualArmRubikCube-v1}, where it achieves 100\% and 84\% success rates, respectively, while both imitation-learning policies achieve at most 8\%. Conversely, it performs poorly on some lower-level tasks, including \texttt{PushSoccer-v1} and \texttt{ShapeTToI-v1}. These results suggest that calibrated difficulty captures a shared aggregate trend, while individual tasks expose complementary strengths and limitations across learning paradigms.

One plausible explanation is that the two imitation-learning policies share a demonstration-driven action-learning formulation, whereas GPT6-Astra constructs task-specific executable controllers through demonstration analysis and online exploration. The pretrained representations of $\pi_{0.5}$ may improve transfer on some tasks, although differences in embodiment and observation processing can constrain these benefits. For GPT6-Astra, explicit control logic and iterative refinement may help organize multi-stage behaviors that are difficult to acquire from a small demonstration set alone. However, errors in visual estimation, controller design, or contact handling may also cause failures on tasks that imitation learning solves reliably.


\section{The \name{} Benchmark}
\label{sec:benchmark}

Following task construction and difficulty calibration, we freeze the task definitions, demonstrations, and evaluation protocol into a versioned release of \name{}.

\subsection{Benchmark Specification}

\name{} contains $112$ tasks grouped into $32$ suites spanning four difficulty levels, two object classes (rigid and deformable), and four embodiments: single-arm gripper (SG), single-arm dexterous hand (SX), dual-arm grippers (DG), and dual-arm dexterous hands (DX). Each suite contains three or four tasks sharing the same difficulty, object class, and embodiment but differing in manipulation skills or objects. Each task has a unique identifier (e.g., \texttt{StackBlock-V1}) and a released configuration specifying its language instruction, initialization distribution, camera setup, action space, control frequency, episode horizon $H_T$, and executable success predicate. An episode succeeds only if the criterion is met within $H_T$ control steps. We additionally provide a scripted expert for demonstration generation.

The standard training set provides $20$ demonstrations per task, totaling $2{,}240$ trajectories at $20$~Hz, with synchronized multi-view RGB observations and action labels. Single-arm tasks use four views (front, left, top, and wrist), while dual-arm tasks use five, with one wrist camera per arm. Images are recorded at $640\times480$, and each task has one language instruction. The standard observation protocol uses RGB images and task instructions only. Proprioception is excluded to maintain a consistent modality across embodiments, while privileged simulator states are withheld because they may directly reveal variables used by the success predicate and are retained only for evaluation. External pretraining is allowed, provided the checkpoint and available training-data information are disclosed.

\subsection{Training and Evaluation Protocols}
\label{sec:benchmark:evaluation}

\paragraph{Training settings}
As \name{} spans multiple dimensions of task variation, we standardize training and evaluation into the following settings:
\begin{itemize}
\item \textbf{Single-Task Setting (ST):} an independent policy is trained and evaluated separately for each task.
\item \textbf{Level-Specific Multi-Task Setting (MT-L):} a single policy is trained jointly on all tasks within each difficulty level and evaluated on that level.
\item \textbf{All-Task Multi-Task Setting (MT-A):} a single policy is trained jointly on the entire benchmark and evaluated across all tasks.
\end{itemize}

For agentic methods, we adopt a D20-E20 setting: each task provides the same 20 training demonstrations used by the imitation-learning baselines, together with a budget of 20 additional environment exploration episodes. During development, the agent may analyze demonstrations, write and revise control programs, and use exploration feedback, while keeping the foundation model and any pretrained tool models frozen. Access is restricted to the task instruction, public robot and control-interface specifications, authorized demonstration observations and actions, and online RGB and proprioceptive observations. Binary success feedback is available during exploration when exposed by the environment interface. Task-specific expert code, hidden simulator states, privileged object information, and evaluation outcomes are prohibited. Every environment trial counts toward the exploration budget, and the native action constraints, episode horizon, and success criteria remain unchanged. After development, the resulting controller and its dependencies are frozen for evaluation; online code modification and weight updates are disallowed. 

\paragraph{Evaluation settings}
Each task is evaluated on exactly $N=50$ test episodes using a fixed set of task-specific initialization seeds shared across methods. Test seeds have been set to be separate from those used for demonstration collection. Let $y_{m,T,i}\in\{0,1\}$ denote the success of method $m$ on task $T$ in test episode $i$. The estimated task success rate is: $\hat p_m(T) = \frac{1}{50}\sum_{i=1}^{50}y_{m,T,i}$. For any multi-task setting $G$, we report the task-averaged success rate, assigning equal weight to each task:$S_m(G) =\frac{1}{|G|}    \sum_{T\in G}\hat p_m(T)$. 

We additionally define \textbf{Attained Level (AL)} as the highest difficulty level at which a method meets a prescribed success criterion on a sufficient fraction of tasks. Let $\mathcal T_k$ denote all tasks in level $k$, $\tau$ the desired task success threshold, and $\rho$ the desired task coverage threshold. The empirical coverage of level $k$ is:
\begin{equation}
    C_k(m;\tau)
    =
    \frac{1}{|\mathcal T_k|}
    \sum_{T\in\mathcal T_k}
    \mathbf 1[\hat p_m(T)\ge\tau].
    \label{eq:benchmark:al_coverage}
\end{equation}
We then define:
\begin{equation}
\begin{aligned}
    \mathrm{AL}_{\tau,\rho}(m)
    =\max\Bigl(&\{0\}\cup
    \bigl\{\ell\in\{1,\ldots,4\}:\\
    &\min_{1\le k\le\ell} C_k(m;\tau)\ge\rho
    \bigr\}\Bigr).
\end{aligned}
\label{eq:benchmark:attained_level}
\end{equation}
Attaining level $\ell$ therefore requires sufficient coverage in that level and every lower level. AL$=0$ indicates that the L1 criterion is unmet. AL can be reported under \textbf{ST} or any \textbf{MT} setting.

\begin{figure*}
    \centering
    \includegraphics[width=\linewidth]{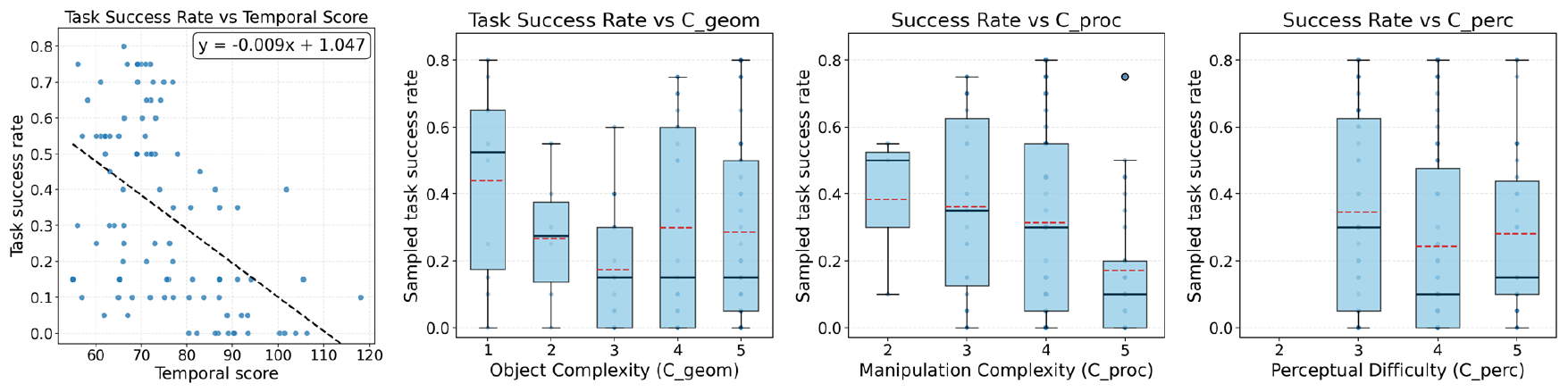}
    \caption{Task success rates with different task properties for all tasks in \name.}
    \vspace{-0.1in}
    \label{fig:stat}
\end{figure*}

In this benchmark, we fix $\rho=0.8$ and report both \textbf{AL@50} ($\tau=0.5$) as the primary level summary, alongside \textbf{AL@80} ($\tau=0.8$) as the strict level summary. With $50$ episodes, these task thresholds correspond to at least $25$ and $40$ observed successes, respectively. Each required level must contain at least $\lceil0.8|\mathcal T_k|\rceil$ qualifying tasks. For example, AL@50$=2$ means that at least $80\%$ of the tasks in each of L1 and L2 achieve an empirical success rate of at least $50\%$, while L3 fails the coverage criterion. Under \textbf{ST}:
\begin{itemize}
    \item \textbf{AL@50} (small flow-matching policy)=1, \textbf{AL@80}  (small flow-matching policy)=0;
    \item \textbf{AL@50} ($\pi_{0.5}$)=1, \textbf{AL@80}  ($\pi_{0.5}$)=0;
    \item \textbf{AL@50} (GPT6-Astra)=0, \textbf{AL@80} (GPT6-Astra)=0;
\end{itemize}

\subsection{Benchmark Analysis and Lessons}

The construction of \name\ provides not only an evaluation platform, but also a large-scale empirical lens for studying what makes manipulation tasks difficult. We annotate each task along four task properties that frequently arise in task taxonomy:

1) Temporal score: the horizon of a task. To reduce the bias introduced by relying solely on demonstration horizon length, e.g., slower execution in some demonstrations, we define the temporal score $T$ of a task as a weighted sum of demonstration steps $N$ and task substages $M$: $T=10*\log N+M$; 

2) Object complexity $C_{\mathrm{geom}}\in\{1,\dots,5\}$, defined by statically inspecting the manipulated objects from tasks, assets, and object metadata. For a task $\mathcal{T}$ with manipulated objects $\mathcal{O}_\mathcal{T}$, we use the most complex task-critical object: $C_{\mathrm{geom}}(\mathcal{T})=\max_{o\in\mathcal{O}_\mathcal{T}} c_{\mathrm{geom}}(o).
$ The object score is rule-based: primitives $=1$, simple rigid household objects $=2$, thin/concave/asymmetric rigid objects $=3$, articulated or multi-part objects $=4$, and deformable or state-dependent objects $=5$;

3) Manipulation complexity $C_{\mathrm{proc}}(\mathcal{T}) = 1+B_{\mathrm{stage}}+B_{\mathrm{contact}}
+B_{\mathrm{coord}}+B_{\mathrm{prec}}, C_{\mathrm{proc}}\in\{1,\dots,5\}
$, where each binary term indicates whether the task requires multi-stage skill
composition, sustained/constrained contact, multi-effector or tool-object
coordination, and strict success tolerance, respectively, annotated from task and robot descriptions, success
conditions, and coded thresholds; 

4) Perception difficulty $C_{\mathrm{perc}}(\mathcal{T})=1+B_{\mathrm{size}}+B_{\mathrm{clutter}}+B_{\mathrm{occlusion}}+B_{\mathrm{hidden}},
\quad
B_i\in\{0,1\}.$
Here, $B_{\mathrm{size}}$ indicates small or thin task-critical objects/contact regions, $B_{\mathrm{clutter}}$ indicates distractors or crowded scenes, $B_{\mathrm{occlusion}}$ indicates partially occluded targets or interaction regions, and $B_{\mathrm{hidden}}$ indicates visually hidden or ambiguous states such as deformable shape or internal container states. We assign these indicators by inspecting assets, scene layout, camera setup, and success conditions.

Figure~\ref{fig:stat} shows that task properties are generally predictive of reference-policy performance. The temporal score is negatively correlated with success rate, indicating that longer-horizon tasks are more difficult due to accumulated execution errors. Higher manipulation complexity and perception difficulty also tend to reduce success rates, suggesting that contact-rich procedures, coordination, strict tolerances, occlusion, and hidden states are key sources of difficulty. The variance further shows that task difficulty arises from multiple factors rather than any single property alone.

\section{Experiments}
  \begin{table*}
      \centering
      \caption{Success rates of $\pi_{0.5}$~\cite{pi05} and changes
  relative to the small flow-matching policy on each task suite in \name.
  Each suite score is the unweighted mean over its tasks; changes in
  parentheses are percentage-point differences. We use $\mathcal{G}$ for
  grippers, $\mathcal{X}$ for dexterous hands, $\mathcal{S}$ for single-
  arm robots, and $\mathcal{D}$ for dual-arm robots.}
      \label{tab:pi05_dp_suite_results}
      \resizebox{\textwidth}{!}{
      \begin{tabular}{c|cccc|cccc}
          \toprule
          & \multicolumn{4}{c|}{Rigid-Body Manipulation}
          & \multicolumn{4}{c}{Deformable-Object Manipulation} \\
          \cmidrule(lr){2-5} \cmidrule(lr){6-9}
          & $\mathcal{SG}$ & $\mathcal{SX}$
          & $\mathcal{DG}$ & $\mathcal{DX}$
          & $\mathcal{SG}$ & $\mathcal{SX}$
          & $\mathcal{DG}$ & $\mathcal{DX}$ \\
          \midrule
          Level 1
          & 48.0\%(\textcolor{red}{\(\downarrow\)23.5\%})
          & 95.5\%(\textcolor{green}{\(\uparrow\)27.5\%})
          & 83.0\%(\textcolor{green}{\(\uparrow\)9.5\%})
          & 59.5\%(\textcolor{red}{\(\downarrow\)13.5\%})
          & 70.7\%(\textcolor{green}{\(\uparrow\)0.7\%})
          & 92.7\%(\textcolor{green}{\(\uparrow\)16.7\%})
          & 68.0\%(\textcolor{red}{\(\downarrow\)2.0\%})
          & 90.7\%(\textcolor{green}{\(\uparrow\)20.7\%}) \\

          Level 2
          & 38.0\%(\textcolor{green}{\(\uparrow\)2.5\%})
          & 52.5\%(\textcolor{green}{\(\uparrow\)17.0\%})
          & 41.5\%(\textcolor{green}{\(\uparrow\)2.5\%})
          & 13.5\%(\textcolor{red}{\(\downarrow\)27.5\%})
          & 50.7\%(\textcolor{green}{\(\uparrow\)13.3\%})
          & 75.3\%(\textcolor{green}{\(\uparrow\)36.0\%})
          & 30.7\%(\textcolor{red}{\(\downarrow\)2.0\%})
          & 28.7\%(\textcolor{red}{\(\downarrow\)10.0\%}) \\
 
          Level 3
          & 11.5\%(\textcolor{green}{\(\uparrow\)3.0\%})
          & 9.0\%(0.0\%)
          & 3.5\%(\textcolor{red}{\(\downarrow\)5.0\%})
          & 6.0\%(\textcolor{red}{\(\downarrow\)3.5\%})
          & 48.7\%(\textcolor{green}{\(\uparrow\)34.0\%})
          & 28.0\%(\textcolor{green}{\(\uparrow\)11.3\%})
          & 17.3\%(\textcolor{green}{\(\uparrow\)4.0\%})
          & 12.7\%(\textcolor{red}{\(\downarrow\)6.0\%}) \\

          Level 4
          & 5.0\%(\textcolor{green}{\(\uparrow\)5.0\%})
          & 2.0\%(\textcolor{green}{\(\uparrow\)2.0\%})
          & 0.5\%(\textcolor{green}{\(\uparrow\)0.5\%})
          & 2.0\%(\textcolor{green}{\(\uparrow\)2.0\%})
          & 2.7\%(\textcolor{green}{\(\uparrow\)2.7\%})
          & 2.0\%(\textcolor{green}{\(\uparrow\)0.7\%})
          & 0.7\%(\textcolor{green}{\(\uparrow\)0.7\%})
          & 1.3\%(\textcolor{green}{\(\uparrow\)1.3\%}) \\
          \bottomrule
      \end{tabular}}
  \end{table*}

\begin{table*}
    \centering
    \caption{Success rates of GPT6-Astra and changes relative to $\pi_{0.5}$ on each task suite in \name. Each suite score is the unweighted mean over its tasks; changes in parentheses are percentage-point differences. We use $\mathcal{G}$ for grippers, $\mathcal{X}$ for dexterous hands, $\mathcal{S}$ for single-arm robots, and $\mathcal{D}$ for dual-arm robots.}
    \label{tab:astra_pi05_suite_results}
    \resizebox{\textwidth}{!}{
    \begin{tabular}{c|cccc|cccc}
        \toprule
        & \multicolumn{4}{c|}{Rigid-Body Manipulation}
        & \multicolumn{4}{c}{Deformable-Object Manipulation} \\
        \cmidrule(lr){2-5} \cmidrule(lr){6-9}
        & $\mathcal{SG}$ & $\mathcal{SX}$
        & $\mathcal{DG}$ & $\mathcal{DX}$
        & $\mathcal{SG}$ & $\mathcal{SX}$
        & $\mathcal{DG}$ & $\mathcal{DX}$ \\
        \midrule
        Level 1
        & 55.5\%(\textcolor{green}{\(\uparrow\)7.5\%})
        & 78.0\%(\textcolor{red}{\(\downarrow\)17.5\%})
        & 83.5\%(\textcolor{green}{\(\uparrow\)0.5\%})
        & 33.0\%(\textcolor{red}{\(\downarrow\)26.5\%})
        & 54.7\%(\textcolor{red}{\(\downarrow\)16.0\%})
        & 95.3\%(\textcolor{green}{\(\uparrow\)2.7\%})
        & 69.3\%(\textcolor{green}{\(\uparrow\)1.3\%})
        & 76.7\%(\textcolor{red}{\(\downarrow\)14.0\%}) \\

        Level 2
        & 44.5\%(\textcolor{green}{\(\uparrow\)6.5\%})
        & 65.5\%(\textcolor{green}{\(\uparrow\)13.0\%})
        & 25.0\%(\textcolor{red}{\(\downarrow\)16.5\%})
        & 51.5\%(\textcolor{green}{\(\uparrow\)38.0\%})
        & 30.7\%(\textcolor{red}{\(\downarrow\)20.0\%})
        & 49.3\%(\textcolor{red}{\(\downarrow\)26.0\%})
        & 94.0\%(\textcolor{green}{\(\uparrow\)63.3\%})
        & 8.0\%(\textcolor{red}{\(\downarrow\)20.7\%}) \\

        Level 3
        & 41.5\%(\textcolor{green}{\(\uparrow\)30.0\%})
        & 27.0\%(\textcolor{green}{\(\uparrow\)18.0\%})
        & 10.5\%(\textcolor{green}{\(\uparrow\)7.0\%})
        & 37.3\%(\textcolor{green}{\(\uparrow\)31.3\%})
        & 37.3\%(\textcolor{red}{\(\downarrow\)11.3\%})
        & 34.7\%(\textcolor{green}{\(\uparrow\)6.7\%})
        & 14.6\%(\textcolor{red}{\(\downarrow\)2.7\%})
        & 0.0\%(\textcolor{red}{\(\downarrow\)12.7\%}) \\

        Level 4
        & 0.0\%(\textcolor{red}{\(\downarrow\)5.0\%})
        & 7.0\%(\textcolor{green}{\(\uparrow\)5.0\%})
        & 38.0\%(\textcolor{green}{\(\uparrow\)37.5\%})
        & 0.0\%(\textcolor{red}{\(\downarrow\)2.0\%})
        & 3.3\%(\textcolor{green}{\(\uparrow\)0.7\%})
        & 13.3\%(\textcolor{green}{\(\uparrow\)11.3\%})
        & 0.0\%(\textcolor{red}{\(\downarrow\)0.7\%})
        & 15.3\%(\textcolor{green}{\(\uparrow\)14.0\%}) \\
        \bottomrule
    \end{tabular}}
\end{table*}

In this section, we conduct experiments to demonstrate how \name\ can be used to evaluate and analyze robot learning methods from multiple perspectives: \textbf{Q1}: How do the success rates of the small flow-matching policy, $\pi_{0.5}$, and GPT6-Astra script policy compare across the various suites? \textbf{Q2}: How does policy performance transfer across different task difficulty levels and task suites? \textbf{Q3}: What is the difference in overall success rate between training each task individually (\textbf{ST}) and training all tasks jointly in a multi-task setting (\textbf{MT-A})? \textbf{Q4}: How does policy performance scale with the amount of demonstration data across different task suites? And \textbf{Q5}: whether the difficulty order calibrated in simulation is preserved in the real world.

For \textbf{Q1}, we show the average task success rates of each task suite in Table~\ref{tab:pi05_dp_suite_results} and Table~\ref{tab:astra_pi05_suite_results}. Compared with the small flow-matching policy, $\pi_{0.5}$ improves on several suites, including Level-1 rigid-body single-arm dexterous tasks (+27.5 percentage points), Level-2 deformable single-arm dexterous tasks (+36.0), and Level-3 deformable single-arm gripper tasks (+34.0). However, these improvements are not uniform: performance decreases on Level-1 rigid-body single-arm gripper tasks (-23.5) and Level-2 rigid-body dual-arm dexterous tasks (-27.5). Thus, pretrained VLA representations provide benefits that depend strongly on the task suite, rather than consistently outperforming a smaller policy trained on the same demonstrations. These results, which do not follow a uniform pattern, confirm the conclusions of some previous studies~\cite{vlaos}: there is no significant difference in success rates between large VLA models and small models when fine-tuned for single tasks.

GPT6-Astra exhibits a different pattern relative to $\pi_{0.5}$. It improves performance across all four Level-3 rigid-body suites, with gains ranging from 7.0 to 31.3 percentage points, and achieves particularly large gains on Level-2 deformable dual-arm gripper tasks (+63.3) and Level-4 rigid-body dual-arm gripper tasks (+37.5). Nevertheless, it underperforms on several lower-level suites and achieves zero success on Level-3 deformable dual-arm dexterous tasks. These results suggest that program construction and iterative exploration can help overcome selected manipulation challenges, while reliable perception and control remain task-dependent bottlenecks. This comparison reflects different learning settings: GPT6-Astra receives additional exploration episodes and proprioceptive observations, so its gains cannot be attributed solely to model capacity.

\finding{No method dominates on \name: $\pi_{0.5}$, GPT6-Astra, and a small flow-matching policy each has its its own advantages on different task suites. The gains from model/method are strongly suite-dependent.}


\begin{table}[t]
    \centering
    \caption{Transfer across difficulty levels with $\pi_{0.5}$. $\mathcal{A}\rightarrow\mathcal{B}$ denotes we first finetune $\pi_{0.5}$ on all tasks of level $\mathcal{A}$, then finetune and evaluate it on all tasks of level $\mathcal{B}$; \emph{Isolated} trains on level $\mathcal{B}$ only. Scores are means over tasks.}
    \label{tab:transfer_level}
    \resizebox{\linewidth}{!}{
    \begin{tabular}{c|c|c|c}
        \toprule
        Setting & All Tasks & Rigid-Body & Deformable \\
        \midrule
        Isolated L1
        & 57.9\% & 57.5\% & 58.5\% \\
        L2 $\rightarrow$ L1
        & 62.0\%(\textcolor{green}{\(\uparrow\)4.1\%})
        & 59.3\%(\textcolor{green}{\(\uparrow\)1.8\%})
        & 65.7\%(\textcolor{green}{\(\uparrow\)7.2\%}) \\
        \midrule
        Isolated L2
        & 24.1\% & 19.8\% & 30.0\% \\
        L1 $\rightarrow$ L2
        & 21.8\%(\textcolor{red}{\(\downarrow\)2.3\%})
        & 20.4\%(\textcolor{green}{\(\uparrow\)0.6\%})
        & 23.7\%(\textcolor{red}{\(\downarrow\)6.3\%}) \\
        L3 $\rightarrow$ L2
        & 19.7\%(\textcolor{red}{\(\downarrow\)4.4\%})
        & 19.0\%(\textcolor{red}{\(\downarrow\)0.8\%})
        & 20.7\%(\textcolor{red}{\(\downarrow\)9.3\%}) \\
        \midrule
        Isolated L3
        & 0.5\% & 0.3\% & 1.0\% \\
        L2 $\rightarrow$ L3
        & 0.2\%(\textcolor{red}{\(\downarrow\)0.3\%})
        & 0.0\%(\textcolor{red}{\(\downarrow\)0.3\%})
        & 0.6\%(\textcolor{red}{\(\downarrow\)0.4\%}) \\
        \bottomrule
    \end{tabular}}
\end{table}

\begin{table}[t]
    \centering
    \caption{Transfer across task suites within Level 2 with $\pi_{0.5}$. $\mathcal{A}\rightarrow\mathcal{B}$ denotes that we first finetune $\pi_{0.5}$ on all tasks of suite $\mathcal{A}$, then finetune and evaluate it on suite $\mathcal{B}$; \emph{Isolated} trains on suite $\mathcal{B}$ only.}
    \label{tab:transfer_suite}
    \resizebox{\linewidth}{!}{
    \begin{tabular}{c|c|c}
        \toprule
        Setting & Isolated & Transfer \\
        \midrule
        Deformable $\rightarrow$ Rigid-Body
        & 17.3\% & 19.9\%(\textcolor{green}{\(\uparrow\)2.6\%}) \\
        Rigid-Body $\rightarrow$ Deformable
        & 16.7\% & 25.8\%(\textcolor{green}{\(\uparrow\)9.2\%}) \\
        \midrule
        Dexterous $ \rightarrow$ Gripper
        & 16.6\% & 22.7\%(\textcolor{green}{\(\uparrow\)6.1\%}) \\
        Gripper $ \rightarrow$ Dexterous
        & 21.0\% & 28.0\%(\textcolor{green}{\(\uparrow\)7.0\%}) \\
        \midrule
        Dual-Arm $ \rightarrow$ Single-Arm
        & 24.1\% & 24.0\%(\textcolor{red}{\(\downarrow\)0.1\%}) \\
        Single-Arm $ \rightarrow$ Dual-Arm
        & 12.0\% & 13.9\%(\textcolor{green}{\(\uparrow\)1.9\%}) \\
        \bottomrule
    \end{tabular}}
\end{table}

For \textbf{Q2}, we test transfer with $\pi_{0.5}$: in a setting $\mathcal{A}\rightarrow\mathcal{B}$, a multi-task LoRA policy is first trained on all tasks of the source set $\mathcal{A}$, then fine-tuned and evaluated on the target set $\mathcal{B}$, and compared with isolated training on $\mathcal{B}$ alone. We consider transfer between Levels 1--3 and between complementary Level-2 suites split by object type, end-effector, and arm setup, using 20 demonstrations and 50 evaluation episodes per task. As shown in Table~\ref{tab:transfer_level}, only L2$\rightarrow$L1 improves the target level (+4.1 percentage points), whereas L1$\rightarrow$L2 and L3$\rightarrow$L2 reduce Level-2 success ($-2.3$ and $-4.4$), mostly on deformable tasks, and Level-2 pretraining leaves Level 3 unsolved. In contrast, Table~\ref{tab:transfer_suite} shows positive transfer in five of the six suite settings, with the largest gains between end-effectors (+6.1 and +7.0) and from rigid-body to deformable tasks (+9.2). Tasks within a level thus share reusable skills, while each harder level requires new capabilities that warm-starting cannot provide.

\finding{Transfer across difficulty levels gives little benefit and is often harmful, whereas transfer between task suites within the same level is mostly beneficial.}

\begin{table*}
    \centering
    \caption{Success rates of $\pi_{0.5}$ trained jointly on all 112 tasks of \name\ (\textbf{MT-A}) and changes relative to single-task training (\textbf{ST}, Table~\ref{tab:pi05_dp_suite_results}) on each task suite. Each suite score is the mean over its tasks. We use $\mathcal{G}$ for grippers, $\mathcal{X}$ for dexterous hands, $\mathcal{S}$ for single-arm robots, and $\mathcal{D}$ for dual-arm robots.}
    \label{tab:mta_suite_results}
    \resizebox{\textwidth}{!}{
    \begin{tabular}{c|cccc|cccc}
        \toprule
        & \multicolumn{4}{c|}{Rigid-Body Manipulation}
        & \multicolumn{4}{c}{Deformable-Object Manipulation} \\
        \cmidrule(lr){2-5} \cmidrule(lr){6-9}
        & $\mathcal{SG}$ & $\mathcal{SX}$
        & $\mathcal{DG}$ & $\mathcal{DX}$
        & $\mathcal{SG}$ & $\mathcal{SX}$
        & $\mathcal{DG}$ & $\mathcal{DX}$ \\
        \midrule
        Level 1
        & 15.5\%(\textcolor{red}{\(\downarrow\)32.5\%})
        & 44.0\%(\textcolor{red}{\(\downarrow\)51.5\%})
        & 16.5\%(\textcolor{red}{\(\downarrow\)66.5\%})
        & 29.0\%(\textcolor{red}{\(\downarrow\)30.5\%})
        & 12.0\%(\textcolor{red}{\(\downarrow\)58.7\%})
        & 36.7\%(\textcolor{red}{\(\downarrow\)56.0\%})
        & 15.3\%(\textcolor{red}{\(\downarrow\)52.7\%})
        & 37.3\%(\textcolor{red}{\(\downarrow\)53.4\%}) \\

        Level 2
        & 10.0\%(\textcolor{red}{\(\downarrow\)28.0\%})
        & 18.0\%(\textcolor{red}{\(\downarrow\)34.5\%})
        & 8.0\%(\textcolor{red}{\(\downarrow\)33.5\%})
        & 1.0\%(\textcolor{red}{\(\downarrow\)12.5\%})
        & 4.7\%(\textcolor{red}{\(\downarrow\)46.0\%})
        & 7.3\%(\textcolor{red}{\(\downarrow\)68.0\%})
        & 3.3\%(\textcolor{red}{\(\downarrow\)27.4\%})
        & 12.0\%(\textcolor{red}{\(\downarrow\)16.7\%}) \\

        Level 3
        & 0.0\%(\textcolor{red}{\(\downarrow\)11.5\%})
        & 0.0\%(\textcolor{red}{\(\downarrow\)9.0\%})
        & 0.0\%(\textcolor{red}{\(\downarrow\)3.5\%})
        & 0.0\%(\textcolor{red}{\(\downarrow\)6.0\%})
        & 0.7\%(\textcolor{red}{\(\downarrow\)48.0\%})
        & 0.0\%(\textcolor{red}{\(\downarrow\)28.0\%})
        & 0.0\%(\textcolor{red}{\(\downarrow\)17.3\%})
        & 0.0\%(\textcolor{red}{\(\downarrow\)12.7\%}) \\

        Level 4
        & 0.0\%(\textcolor{red}{\(\downarrow\)5.0\%})
        & 1.0\%(\textcolor{red}{\(\downarrow\)1.0\%})
        & 0.0\%(\textcolor{red}{\(\downarrow\)0.5\%})
        & 0.0\%(\textcolor{red}{\(\downarrow\)2.0\%})
        & 0.0\%(\textcolor{red}{\(\downarrow\)2.7\%})
        & 0.0\%(\textcolor{red}{\(\downarrow\)2.0\%})
        & 0.0\%(\textcolor{red}{\(\downarrow\)0.7\%})
        & 0.0\%(\textcolor{red}{\(\downarrow\)1.3\%}) \\
        \bottomrule
    \end{tabular}}
\end{table*}

\begin{figure*}
    \centering
    \includegraphics[width=\linewidth]{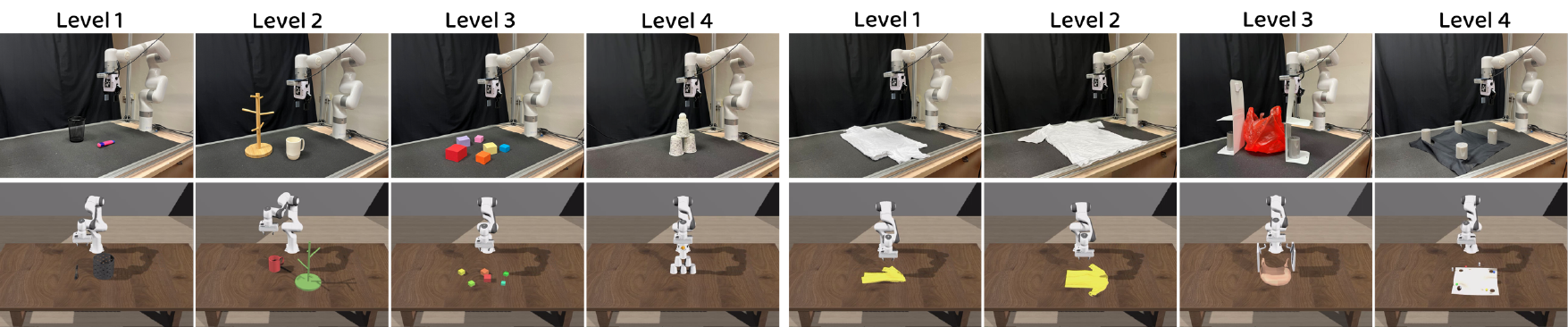}
    \caption{We choose 8 tasks from four different task difficulty levels for real-world experiments: \textit{PickPlacePen}, \textit{HangMug}, \textit{Stack6Blocks}, \textit{PingPongTransport}, \textit{UnfoldShirt}, \textit{FoldShirt}, \textit{BagHanging}, and \textit{BuildTent}, with corresponding simulation tasks at the bottom. Results are in Table \ref{tab:real_sim_success}.}
    \label{fig:real_world}
\end{figure*}

\begin{table*}[t]
    \centering
    \caption{Comparison between real-world and the corresponding simulation task success rates.}
    \label{tab:real_sim_success}
    \resizebox{\linewidth}{!}{
    \begin{tabular}{lcccccccc}
        \toprule
        \textbf{Task} 
        & \textbf{PickPlacePen} 
        & \textbf{HangMug} 
        & \textbf{Stack6Blocks} 
        & \textbf{PingPongTransport} 
        & \textbf{UnfoldShirt} 
        & \textbf{FoldShirt} 
        & \textbf{BagHanging} 
        & \textbf{BuildTent} \\
        \midrule
        \textbf{Real-world}
        & 100\% & 40\% & 10\% & 0\% & 80\% & 40\% & 10\% & 0\% \\
        \textbf{Simulation}
        & 80\% & 44\% & 10\% & 0\% & 70\% & 44\% & 6\% & 0\% \\
        \bottomrule
    \end{tabular}
    }
    \vspace{-0.2in}
\end{table*}

For \textbf{Q3}, we jointly fine-tune a single $\pi_{0.5}$ policy on all 112 tasks of \name\ (\textbf{MT-A}) and compare it with single-task training (\textbf{ST}). MT-A uses the same 20 demonstrations per task and is trained for 112$\times$ steps in total than the single task training budget; we evaluate 50 episodes per task. As shown in Table~\ref{tab:mta_suite_results}, MT-A is worse than ST on all 32 suites. Its average success rate drops from 75.4\% to 25.9\% on Level 1 and from 40.7\% to 8.2\% on Level 2, and it almost never succeeds on Levels 3 and 4 (0.1\% each), giving 8.6\% over all tasks versus 33.5\% for ST. Training one policy per level (the isolated rows of Table~\ref{tab:transfer_level}) falls in between, with 57.9\% and 24.1\% on Levels 1 and 2, so success degrades steadily as more tasks share one policy.

\finding{A single policy trained jointly on all 112 tasks is far worse than single-task policies on every suite (8.6\% vs.\ 33.5\% overall): multi-task learning at benchmark scale remains an open challenge.}

\begin{figure}
    \centering
    \includegraphics[width=\linewidth]{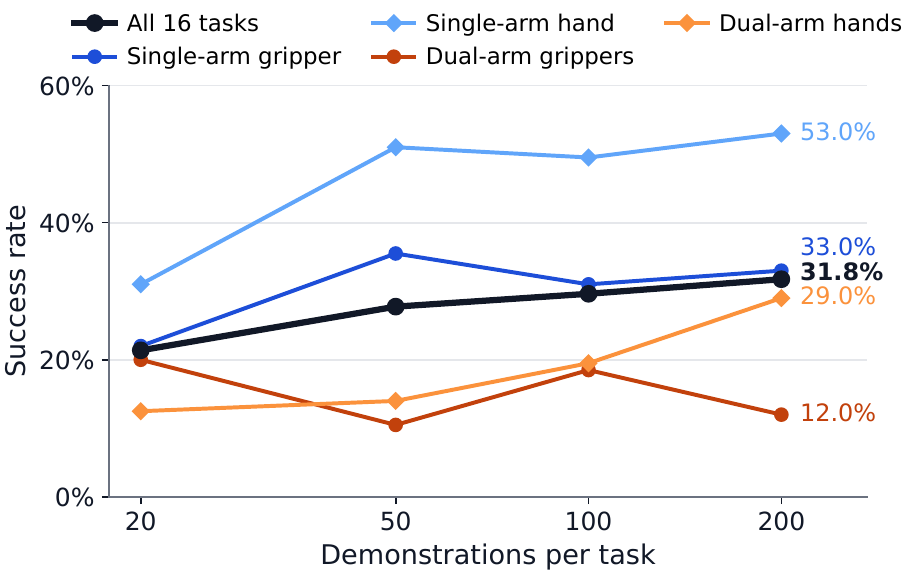}
    \caption{Data scaling of $\pi_{0.5}$ on the 16 Level-2 rigid-body tasks. One policy is trained jointly on all tasks with 20--200 demonstrations per task. The black curve is the mean success rate over all 16 tasks; colored curves are suite means (4 tasks, 50 episodes each).}
    \vspace{-0.15in}
    \label{fig:data_scaling}
\end{figure}

For \textbf{Q4}, we study whether \name\ is saturated by data. We jointly fine-tune $\pi_{0.5}$ on the 16 Level-2 rigid-body tasks with 20, 50, 100, or 200 demonstrations per task for 100k steps and evaluate 50 episodes per task. As shown in Figure~\ref{fig:data_scaling}, the average success rate increases monotonically with the amount of data, from 21.4\% to 27.8\%, 29.6\%, and 31.8\%, so the tasks are learnable and additional demonstrations keep helping. The improvement continues up to 200 demonstrations, most visibly for the dexterous-hand suites ($\mathcal{SX}$: 31.0\%$\rightarrow$53.0\%; $\mathcal{DX}$: 12.5\%$\rightarrow$29.0\%). At the same time, ten times more data still leaves the average below 32\%, and dual-arm gripper tasks do not benefit at all (20.0\%$\rightarrow$12.0\%). \name\ is therefore far from saturated in the evaluated data regime: more data brings steady gains, while large headroom remains even at Level 2.

\finding{\name\ is not saturated by data: success rates keep rising as demonstrations grow from 20 to 200, yet ten times more data still leaves Level-2 rigid-body tasks below 32\%, with ample headroom for better methods.}

For \textbf{Q5}, we select 8 representative tasks from the four difficulty levels of \name\, covering both rigid-body and deformable-object manipulation. For each task, we construct a corresponding real-world setup with similar object geometry, goal condition, and manipulation procedure as its simulation counterpart. We use the flow-matching policy and 20 demonstrations for each task. We evaluate each real-world task for 50 trials and compare the results with the success rate of the corresponding simulation task. We use X-Arm as the robot arm, use Meta Quest 3 as the real-world teleoperation tool, and use two RealSense D435i cameras, as shown in Figure~\ref{fig:real_world}. As in Table~\ref{tab:real_sim_success}, the real-world results largely preserve the difficulty ordering calibrated in simulation. 

\finding{The difficulty ordering calibrated in simulation carries over to the real world: real-world success rates drop level by level and track their simulation counterparts.}

\section{Conclusion}
We introduced \name, a difficulty-calibrated simulation benchmark for robotic manipulation that organizes 112 tasks across 4 difficulty levels and diverse task categories. By calibrating task difficulty with the success rate of a standardized imitation-learning policy and iteratively refining task designs, \name\ provides a policy-anchored ladder for measuring what level of manipulation difficulty the policy can reliably solve. We demonstrate that \name\ can support various studies. We hope \name\ will serve as a structured testbed for understanding the boundaries of robot learning methods and guiding future progress toward more capable manipulation systems.


{\small
\bibliographystyle{IEEEtran}
\bibliography{references}
}


\end{document}